\documentclass[runningheads]{llncs}
\usepackage{graphicx}
\usepackage{amsmath}
\usepackage{algorithm}
\usepackage{algpseudocode}
\usepackage{amssymb}
\usepackage{xspace}
\usepackage{xcolor}
\usepackage{bm}
\usepackage{array}
\usepackage{hyperref}
\hypersetup{
	colorlinks=true,
	urlcolor=blue,
}

\def\eg{\emph{e.g.}\xspace} 
\def\ie{\emph{i.e.}\xspace}

\newcommand{\figref}[1]{{Figure~\ref{#1}}}
\newcommand{\Figref}[1]{{Figure~\ref{#1}}}
\newcommand{\secref}[1]{{Section~\ref{#1}}}

\renewcommand{\eqref}[1]{{Equation~(\ref{#1})}}

\newcommand{\qt}[1]{`{#1}'}
\newcommand{\argmin}[2]{\arg\,\min\limits_{#1} {#2}}
\newcommand{\argmax}[2]{\arg\,\max\limits_{#1} {#2}}
\newcommand{\ind}[2]{{\left[ #1 \right]_{#2}}}

\newcommand{\R}{\mathbb{R}}

\newcolumntype{L}[1]{>{\raggedright\let\newline\\\arraybackslash\hspace{0pt}}m{#1}}
\newcolumntype{C}[1]{>{\centering\let\newline\\\arraybackslash\hspace{0pt}}m{#1}}
\newcolumntype{R}[1]{>{\raggedleft\let\newline\\\arraybackslash\hspace{0pt}}m{#1}}
\newcolumntype{N}{@{}m{0pt}@{}}

\begin{document}
	\title{Cost-efficient segmentation of electron microscopy images using active learning\thanks{We gratefully acknowledge the support of NVIDIA Corporation with the donation of the Titan V GPU used for this research. Y.S. is a Marylou Ingram Scholar. }}
	\author{Joris Roels\inst{1,2}\orcidID{0000-0002-2058-8134} \and
		Yvan Saeys\inst{1,2}\orcidID{0000-0002-0415-1506}}
	\authorrunning{J. Roels et al.}
	\titlerunning{Cost-efficient segmentation of EM images using active learning}
	\institute{
		Department of Applied Mathematics, Computer Science and Statistics, Ghent University, Ghent, Belgium\\
		\email{\{jorisb.roels, yvan.saeys\}@ugent.be}
		\and
		Inflammation Research Center, Flanders Institute for Biotechnology, Ghent, Belgium
	}
	\maketitle              % typeset the header of the contribution
	\begin{abstract}
		Over the last decade, electron microscopy has improved up to a point that generating high quality gigavoxel sized datasets only requires a few hours. Automated image analysis, particularly image segmentation, however, has not evolved at the same pace. Even though state-of-the-art methods such as U-Net and DeepLab have improved segmentation performance substantially, the required amount of labels remains too expensive. Active learning is the subfield in machine learning that aims to mitigate this burden by selecting the samples that require labeling in a smart way. Many techniques have been proposed, particularly for image classification, to increase the steepness of learning curves. In this work, we extend these techniques to deep CNN based image segmentation. Our experiments on three different electron microscopy datasets show that active learning can improve segmentation quality by 10 to 15$\%$ in terms of Jaccard score compared to standard randomized sampling. 
	\end{abstract}
	\section{Introduction}
	
	% image segmentation in electron microscopy
	Semantic image segmentation, the task of assigning pixel-level object labels to an image, is a fundamental task in many applications and one of the most challenging problems in generic computer vision. Particularly in biomedical imaging such as electron microscopy (EM), where annotated data is very sparsely available and image data contains high resolution ($\approx$ 5 nm$^3$) and ultrastructural content. Nevertheless, deep learning has caused significant improvements in this particular research domain, over the last years \cite{Ciresan2012a,Ronneberger2015,Januszewski2018High-precisionNetworks}. 
	
	% the issue of labels (particularly in biomedical imaging)
	Even though the impressive advances that have been made so far, state-of-the-art techniques mostly rely on large annotated datasets. This is an impractical assumption and only satisfied for particular use-cases such as \eg neuron segmentation \cite{Arganda-Carreras2015}. For segmentation of alternative classes, research often falls back to manual segmentation or interactive approaches that rely on shallow segmentation algorithms \cite{Sommer2011,Belevich2016,Arganda-Carreras2017TrainableClassification}, which is costly or sacrifices performance. 
	
	% active learning
	This work focuses on active learning, a subdomain of machine learning that aims to minimize supervision without sacrificing predictive accuracy. This is achieved by iteratively querying a batch of samples to a label providing oracle, adding them to the train set and retraining the predictor. The challenge is to come up with a smart selection criterion to query samples and maximize the steepness of the training curve \cite{Settles2010ActiveSurvey}. 
	
	% proposed work
	In this work, we employ state-of-the-art active learning approaches, commonly used for classification, to image segmentation. Particularly, we illustrate on three EM datasets that the amount of annotated samples can be reduced to a few hundreds to obtain close to fully supervised performance. We start by formally defining the active learning problem in the context of image segmentation in \secref{sec:notations}. In \secref{sec:active-learning}, we give an overview of commonly used, recent active learning approaches in classification \cite{Settles2010ActiveSurvey} and how these techniques can be used in segmentation. This is followed by experimental results and a discussion in \secref{sec:experiments}. Lastly, the paper is concluded in \secref{sec:conclusion}. 
	
	\section{Notations}
	\label{sec:notations}
	We consider the task of image segmentation, \ie given an $N$ pixel image $\vec{x} \in X \subset \R^N$, we aim to compute a pixel-level labeling $\vec{y} \in Y$, where $Y = \{0, \dots, C-1\}^N$ is the label space and $C$ is the number of classes. We particularly focus on the case of binary segmentation, \ie $C=2$. Let $\vec{p}_j(\vec{x}) = \ind{\vec{f}_{\vec{\theta}}(\vec{x})}{j}$ be the probability class distribution of pixel $j$ of a parameterized segmentation algorithm $\vec{f}_{\vec{\theta}}$ (for example, an encoder-decoder network such as U-Net \cite{Ronneberger2015}). 
	
\begin{figure}[t!]
		\centering
		\includegraphics[width=\linewidth]{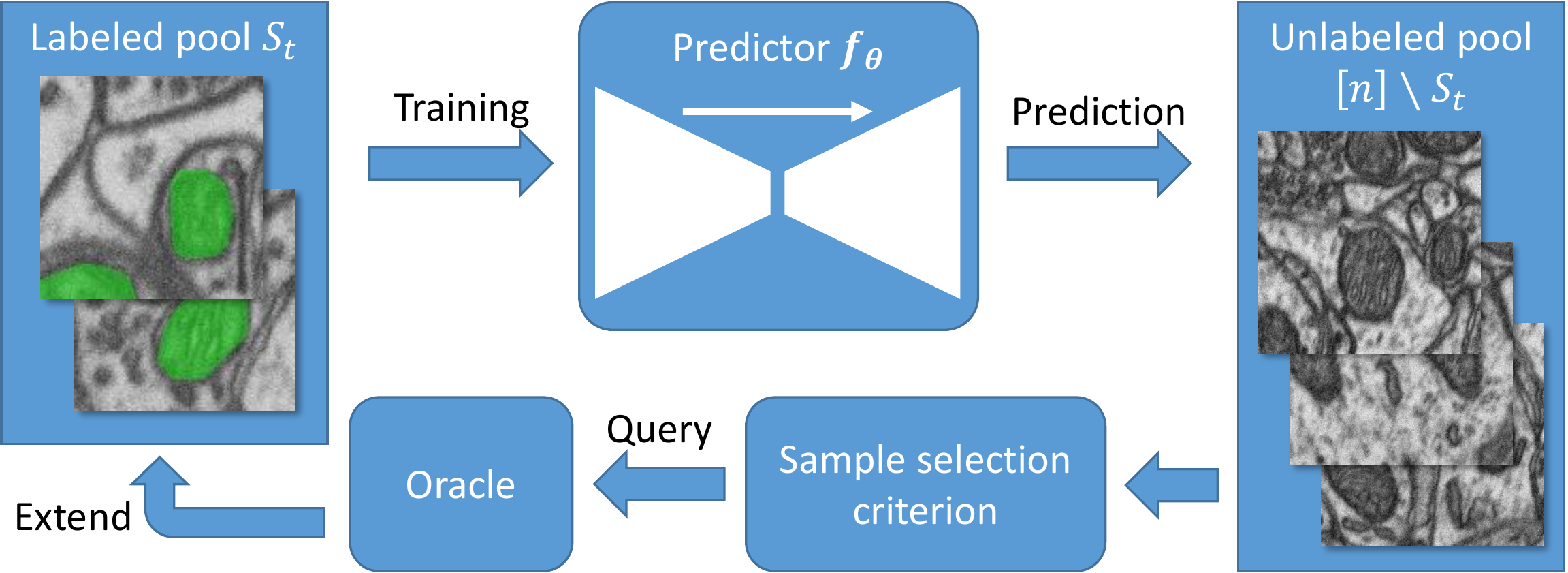} 
		\caption{Iterative active learning workflow for segmentation. A predictor network $\vec{f}_{\vec{\theta}}$ predicts the class probability distributions of the unlabeled samples. These outputs are used in a sample criterion to select the \qt{most informative} samples. The selected samples are labeled by an oracle and the extended labeled pool is used to retrain the predictor. }
		\label{fig:active-learning-workflow}
	\end{figure}

	Consider a large pool of i.i.d. sampled data points over the space $Z = X \times Y$ as $\{ \vec{x}_i, \vec{y}_i \}_{i \in [n]}$, where $[n] = \{1, \dots, n\}$, and an initial pool of $m$ randomly chosen distinct data points indexed by $S_0 = \{i_j | i_j \in [n]\}_{j \in [m]}$. An active learning algorithm initially only has access to $\{ \vec{x}_i \}_{i \in [n]}$ and $\{ \vec{y}_i \}_{i \in S_0}$ and iteratively extends the currently labeled pool $S_t$ by querying $k$ samples from the unlabeled set $\{ \vec{x}_i \}_{i \in [n] \backslash S_t}$ to an oracle. After iteration $t$, the predictor is retrained with the available samples $\{ \vec{x}_i \}_{i \in [n]}$ and labels $\{ \vec{y}_i \}_{i \in S_t}$, thereby improving the segmentation quality. Note that, without loss of generalization, the active learning approaches below are described for $k=1$ as we can also query $k>1$ samples for $k$ iterations, without retraining. The complete active learning workflow is shown in \figref{fig:active-learning-workflow}. 
	
	\section{Active learning}
	\label{sec:active-learning}
	In the following sections, we will discuss 5 well known and recent active learning approaches for classification: maximum entropy selection \cite{Joshi2009Multi-classClassification,Li2013AdaptiveClassification}, least confidence selection \cite{Blinker2003IncorporatingMachines}, Bayesian active learning disagreement \cite{Gal2017DeepData}, k-means sampling \cite{Bodo2011ActiveClustering} and core set active learning \cite{Sener2018ActiveApproach}. Furthermore, we will show how these techniques can be applied to image segmentation. 
	
	% \begin{itemize}
	%     \item Survey classical techniques \cite{Settles2010ActiveSurvey}
	%     \item Uncertainty based: entropy \cite{Joshi2009Multi-classClassification,Li2013AdaptiveClassification}, distance to decision surface \cite{Blinker2003IncorporatingMachines}
	%     \item Bayesian approaches: estimate expected improvements \cite{Kapoor2007ActiveCategorization}, dropout uncertainty \cite{Gal2017DeepData}
	%     \item Diversity enforcing: clustering \cite{Bodo2011ActiveClustering}
	%     \item Uncertainty - diversity tradeoff \cite{Yang2015Multi-ClassMaximization,Guo2010ActivePartition}
	%     \item Core set approach \cite{Sener2018ActiveApproach}
	% \end{itemize}
	
	\subsection{Maximum entropy sampling}
	
	Maximum entropy is a straightforward selection criterion that aims to select samples for which the predictions are uncertain \cite{Joshi2009Multi-classClassification,Li2013AdaptiveClassification}. Formally speaking, we adjust the selection criterion to a pixel-wise entropy calculation as follows: 
	\begin{equation}\label{eq:max-entropy}
	\vec{x}_{t+1}^* = \argmax{\vec{x} \in [n] \backslash S_t}{- \sum\limits_{j=0}^{N-1} \sum\limits_{c=0}^{C-1} \ind{\vec{p}_j(\vec{x})}{c} \log{\ind{\vec{p}_j(\vec{x})}{c}} }.
	\end{equation}
	In other words, the entropy is calculated for each pixel and cumulated. Note that a high entropy will be obtained when $\vec{p}_j(\vec{x}) = \frac{1}{C}$, this is exactly when there is no real consensus on the predicted class (\ie high uncertainty). 
	
	\subsection{Least confidence sampling}
	Similar to maximum entropy sampling, the least confidence criterion selects samples for which the predictions are uncertain: 
	\begin{equation}\label{eq:least-confidence}
	\vec{x}_{t+1}^* = \argmin{\vec{x} \in [n] \backslash S_t}{\sum\limits_{j=0}^{N-1} \max\limits_{c=0,\dots,C-1} \ind{\vec{p}_j(\vec{x})}{c} }.
	\end{equation}
	As the name suggest, the least confidence criterion selects the probability that corresponds to the predicted class. Whenever this probability is small, the predictor is not confident about this decision. For image segmentation, we cumulate the maximum probabilities to select the least confident samples. 
	
	\subsection{Bayesian active learning disagreement}
	The Bayesian active learning disagreement (BALD) approach \cite{Gal2017DeepData} is specifically designed for convolutional neural networks (CNNs). It makes use of Bayesian CNNs in order to cope with the small amounts of training data that are usually available in active learning workflows. A Bayesian CNN assumes a prior probability distribution placed over the model parameters $\vec{\theta} \sim p(\vec{\theta})$. The uncertainty in the weights induces prediction uncertainty by marginalising over the approximate posterior \cite{Gal2017DeepData}: 
	\begin{eqnarray}\label{eq:bald}
	\ind{\vec{p}_j(\vec{x})}{c} \approx \frac{1}{T} \sum\limits_{t=0}^{T-1} \ind{\vec{p}_j(\vec{x} ; \hat{\vec{\theta}}_t)}{c},
	\end{eqnarray}
	where $\hat{\vec{\theta}}_t \sim q(\vec{\theta})$ is the dropout distribution, which approximates the prior probability distribution $p$. In other words, a CNN is trained with dropout and inference is obtained by leaving dropout on. This causes uncertainty in the outcome that can be used in existing criteria such as maximum entropy (\eqref{eq:max-entropy}). 
	
	\subsection{K-means sampling}
	Uncertainty-based approaches typically sample close to the decision boundary of the classifier. This introduces an implicit bias that does not allow for data exploration. Most explorative approaches that aim to solve this problem transform the input $\vec{x}$ to a more compact and efficient representation $\vec{z} = \vec{g}(\vec{x})$ (\eg the feature representation before the fully connected stage in a classification CNN). The representation that we used in our segmentation approach was the bottleneck representation in the U-Net. The $k$-means sampling approach in particular then finds $k$ clusters in this embedding using $k$-means clustering. The selected samples are then the $k$ samples in the different clusters that are closest to the $k$ centroids. 
	
	\subsection{Core set active learning}
	The core set approach \cite{Sener2018ActiveApproach} is a recently proposed active learning approach for CNNs that is not based on uncertainty or exploratory sampling. Similar to $k$-means, samples are selected from an embedding $\vec{z} = \vec{g}(\vec{x})$ in such a way that a model trained on the selection of samples would be competitive for the remaining samples. Similar as before, the representation that we used in our segmentation approach was the bottleneck representation in the U-Net. In order to obtain such competitive samples, this approach aims to minimize the so-called core set loss. This is the difference between average empirical loss over the set of labeled samples (\ie $S_t$) and the average empirical loss over the entire dataset including unlabelled points (\ie $[n]$).

	\section{Experiments \& discussion}
	\label{sec:experiments}
	
	\begin{figure}[t!]
		\centering
		\begin{minipage}{0.48\linewidth}
			\centering
			\includegraphics[width=0.95\linewidth]{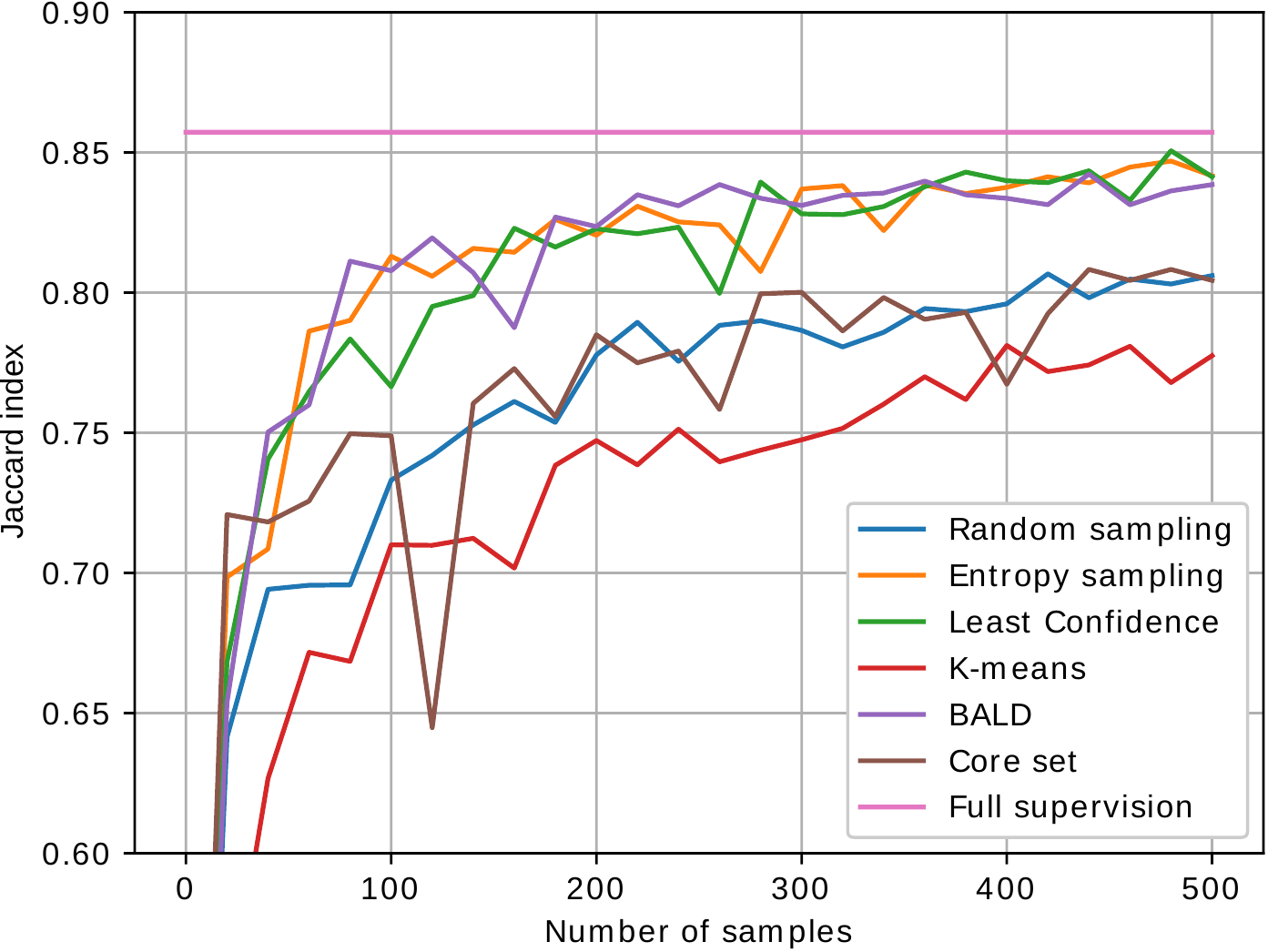} 
			\small\centerline{(a) EPFL}
			\vspace{0.1cm}
		\end{minipage}
		\begin{minipage}{0.48\linewidth}
			\centering
			\includegraphics[width=0.95\linewidth]{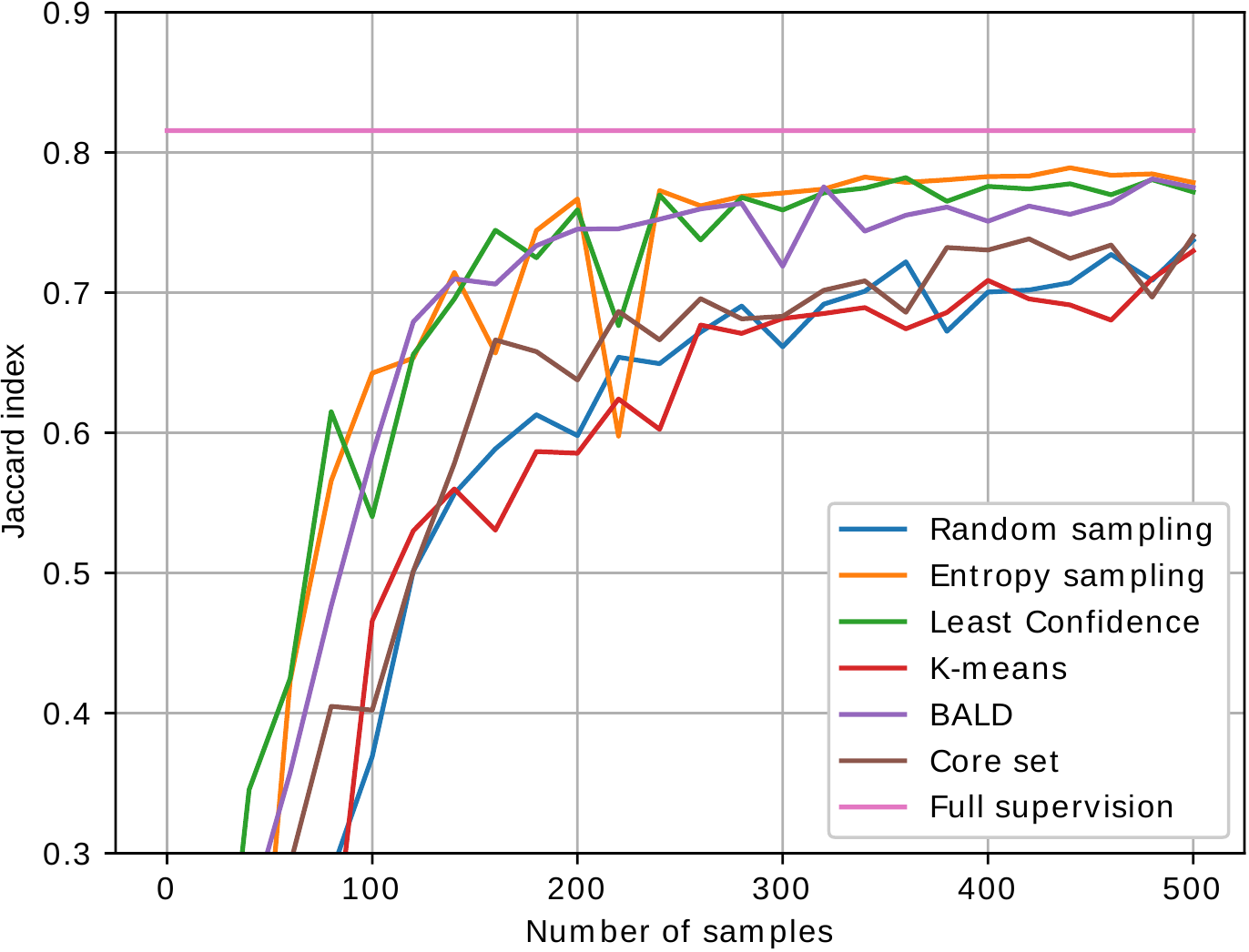} 
			\small\centerline{(b) VNC}
			\vspace{0.1cm}
		\end{minipage}
		\begin{minipage}{0.48\linewidth}
			\centering
			\includegraphics[width=0.95\linewidth]{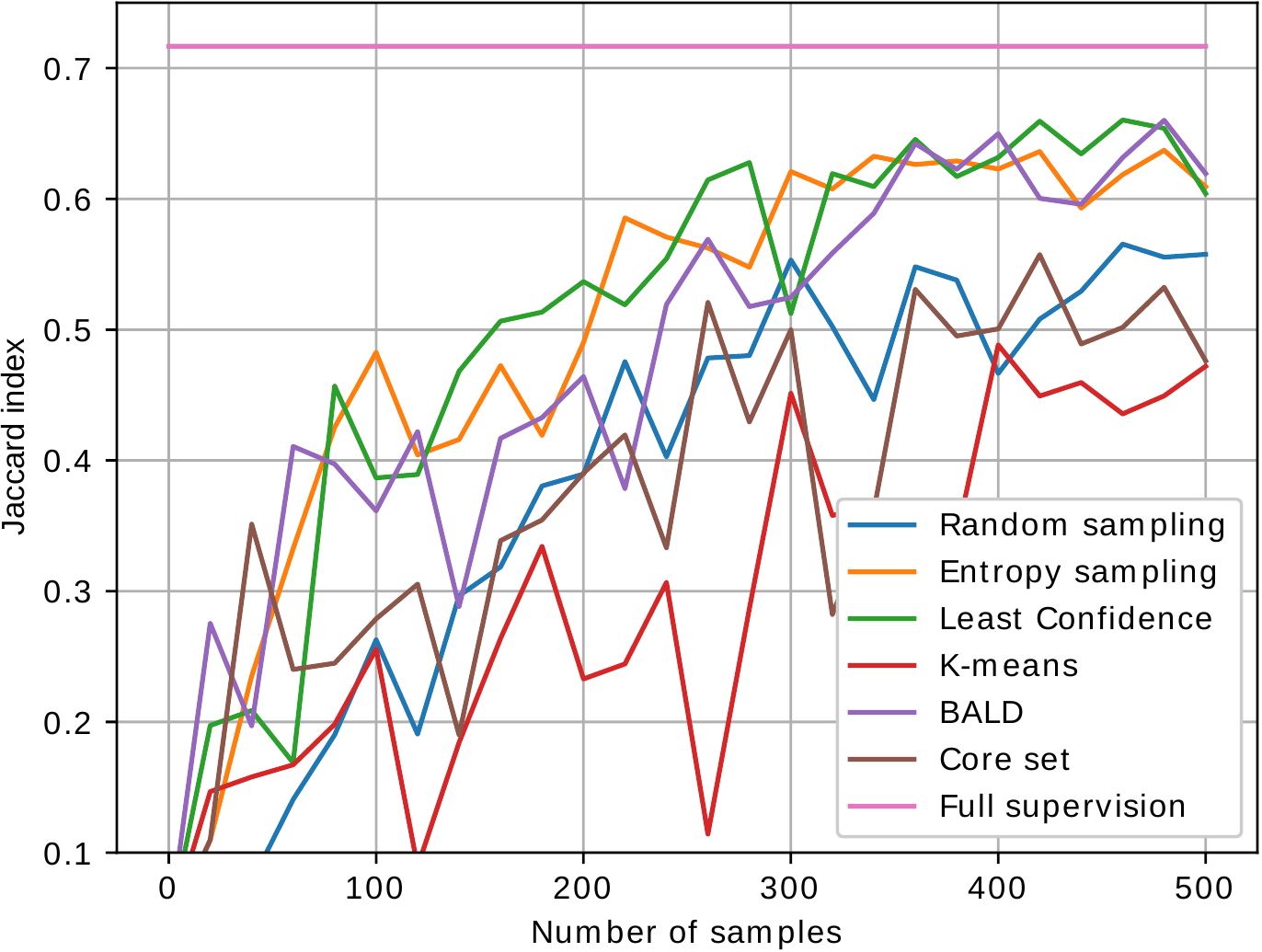} 
			\small\centerline{(c) MiRA}
		\end{minipage}
		\caption{Learning curves for the discussed active learning approaches for the different datasets. }
		\label{fig:learning-curves}
	\end{figure}
	
	Three public EM datasets where used to validate our approach:
	\begin{itemize}
		\item The EPFL dataset\footnote{Data available at \href{https://cvlab.epfl.ch/data/data-em/}{https://cvlab.epfl.ch/data/data-em/}} represents a $5 \times 5 \times 5$ $\mu$m$^3$ section taken from the CA1 hippocampus region of the brain, corresponding to a $2048 \times 1536 \times 1065$ volume. Two $1048 \times 786 \times 165$ subvolumes were manually labeled by experts for mitochondria. The data was acquired by a focused ion-beam scanning EM and the resolution of each voxel is approximately $5 \times 5 \times 5$ nm$^3$. 
		\item The VNC dataset\footnote{Data available at \href{https://github.com/unidesigner/groundtruth-drosophila-vnc/}{https://github.com/unidesigner/groundtruth-drosophila-vnc/}} represents two $4.7 \times 4.7 \times 1$ $\mu$m$^3$ sections taken from the Drosophila melanogaster third instar larva ventral nerve cord, corresponding to a $1024 \times 1024 \times 20$ volume. One stack was manually labeled by experts for mitochondria. The data was acquired by a transmission EM and the resolution of each voxel is approximately $4.6 \times 4.6 \times 45$ nm$^3$. 
		\item The MiRA dataset\footnote{Data available at \href{http://95.163.198.142/MiRA/mitochondria31/}{http://95.163.198.142/MiRA/mitochondria31/}} \cite{Xiao2018a} represents a $17 \times 17 \times 1.6$ $\mu$m$^3$ section taken from the mouse cortex, corresponding to a $8624 \times 8416 \times 31$ volume. The complete volume was manually labeled by experts for mitochondria. The data was acquired by an automated tape-collecting ultramicrotome scanning EM and the resolution of each voxel is approximately $2 \times 2 \times 50$ nm$^3$. 
	\end{itemize}
	
	To properly validate the discussed approaches, we split the available labeled data in a training and testing set. In the cases of a single labeled volume (VNC and MiRA), we split these datasets halfway along the $y$ axis. A smaller U-Net (with 4 times less feature maps) was initially trained on $m=20$ randomly selected $128 \times 128$ samples in the training volume (learning rate of $1e^{-3}$ for 500 epochs). Next, we consider a pool of $n=2000$ samples in the training data to be queried. Each iteration, $k=20$ samples are selected from this pool based on one of the discussed selection criteria, and added to the labeled set $S_t$, after which the segmentation network is finetuned (learning rate of $5e^{-4}$ for 200 epochs). This procedure is repeated for $T=25$ iterations, leading to a maximum training set size of 500 samples. We validate the segmentation performance with the well known Jaccard score: 
	\begin{equation}
	J(\vec{y}, \hat{\vec{y}}) = \frac{\sum_i \ind{\vec{y} \cdot \hat{\vec{y}}}{i}}{\sum_i \ind{\vec{y}}{i} + \sum_i \ind{\hat{\vec{y}}}{i} - \sum_i \ind{\vec{y} \cdot \hat{\vec{y}}}{i}}
	\end{equation}
	This segmentation metric is also known as the intersection-over-union (IoU). 
	
	\begin{figure}[t!]
		\centering
		\begin{minipage}{0.32\linewidth}
			\centering
			\includegraphics[width=\linewidth]{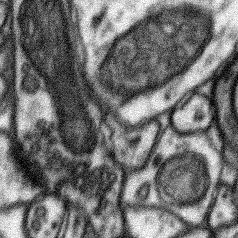} 
			\small\centerline{(a) Input}
		\end{minipage}
		\begin{minipage}{0.32\linewidth}
			\centering
			\includegraphics[width=\linewidth]{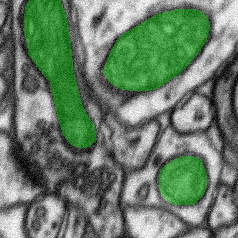} 
			\small\centerline{(b) Ground truth}
		\end{minipage}
		\begin{minipage}{0.32\linewidth}
			\centering
			\includegraphics[width=\linewidth]{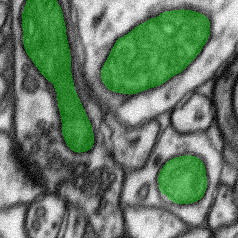} 
			\small\centerline{(c) Full supervision ($0.857$)}
		\end{minipage}
		\begin{minipage}{0.32\linewidth}
			\centering
			\includegraphics[width=\linewidth]{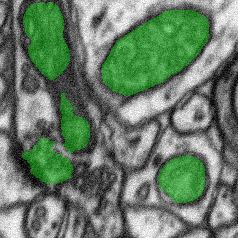} 
			\small\centerline{(d) Random ($0.733$)}
		\end{minipage}
		\begin{minipage}{0.32\linewidth}
			\centering
			\includegraphics[width=\linewidth]{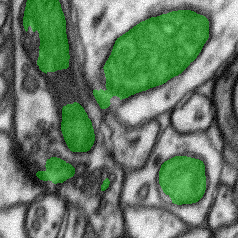} 
			\small\centerline{(e) $k$-means ($0.710$)}
		\end{minipage}
		\begin{minipage}{0.32\linewidth}
			\centering
			\includegraphics[width=\linewidth]{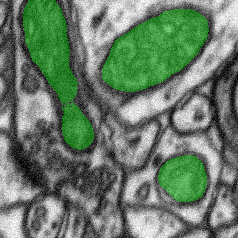} 
			\small\centerline{(f) Maximum entropy ($0.813$)}
		\end{minipage}
		\caption{Segmentation results obtained from an actively learned U-Net with 120 samples of the EPFL dataset based on random, $k$-means and maximum entropy sampling, and a comparison to the fully supervised approach. Jaccard scores are indicated between brackets. }
		\label{fig:segmentation-results}
	\end{figure}
	
	The resulting performance curves of the discussed approaches on the three datasets are shown in \figref{fig:learning-curves}. We additionally show the performance obtained by full supervision (\ie all labels are available during training), which is the maximum achievable segmentation performance. In comparison to the random sampling baseline, we observe that the maximum entropy, least confidence and BALD approach perform significantly better. These methods obtain about 10 to 15$\%$ performance increase for the same amount of available labels for all datasets. The recently proposed core set approach performs similar to slightly better than the baseline. We expect that this method can be improved by considering alternative embeddings. Lastly, we see that $k$-means performs significantly worse than random sampling. Even though this could also be an embedding problem such as with the core set approach, we think that exploratory sampling alone will not allow the predictor to learn from challenging samples, which are usually outliers. We expect that a hybrid approach based on both exploration and uncertainty might lead to better results, and consider this future work. 
	
	\begin{figure}[t!]
		\centering
		\begin{minipage}{0.24\linewidth}
			\centering
			\includegraphics[width=\linewidth]{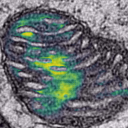} 
		\end{minipage}
		\begin{minipage}{0.24\linewidth}
			\centering
			\includegraphics[width=\linewidth]{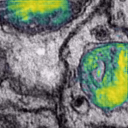} 
		\end{minipage}
		\begin{minipage}{0.24\linewidth}
			\centering
			\includegraphics[width=\linewidth]{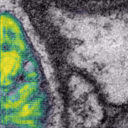} 
		\end{minipage}
		\begin{minipage}{0.24\linewidth}
			\centering
			\includegraphics[width=\linewidth]{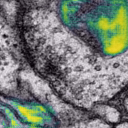} 
		\end{minipage}
		\begin{minipage}{0.24\linewidth}
			\centering
			\includegraphics[width=\linewidth]{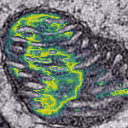} 
			\small\centerline{$t=1$}
		\end{minipage}
		\begin{minipage}{0.24\linewidth}
			\centering
			\includegraphics[width=\linewidth]{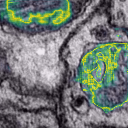} 
			\small\centerline{$t=2$}
		\end{minipage}
		\begin{minipage}{0.24\linewidth}
			\centering
			\includegraphics[width=\linewidth]{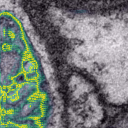} 
			\small\centerline{$t=3$}
		\end{minipage}
		\begin{minipage}{0.24\linewidth}
			\centering
			\includegraphics[width=\linewidth]{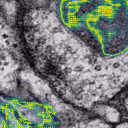} 
			\small\centerline{$t=4$}
		\end{minipage}
		\caption{Illustration of the selected samples in the VNC dataset over time in the active learning process. The top row shows the pixel-wise prediction of the selected samples at iterations 1 through 4. The bottom row show the pixel-wise least confidence score on the corresponding images. }
		\label{fig:selected samples}
	\end{figure}
	
	\Figref{fig:segmentation-results} shows qualitative segmentation results on the EPFL dataset. In particular, we show results of the random, $k$-means and maximum entropy sampling methods using 120 samples, and compare this to the fully supervised approach. The maximum entropy sampling technique is able to improve the others by a large margin and closes the gap towards fully supervised learning significantly. 
	
	Lastly, we are interested in what type of samples the active learning approaches select for training. \Figref{fig:selected samples} shows 4 samples of the VNC dataset that correspond to the highest prioritized samples, according to the least confidence criterion, that were selected in the first 4 iterations. The top row illustrates the probability predictions of the network at that point in time, whereas the bottom row shows the pixel-wise uncertainty of the sample (\ie the maximum in \eqref{eq:least-confidence}). Note that the initial predictions at $t=1$ are of poor quality, as the network was only trained on 20 samples. Moreover, the uncertainty is high in regions where the network is uncertain, but it is low in regions where the network is wrong. The latter is a common issue in active learning and related to the exploration vs. uncertainty trade off. However, over time, we see that the network performance improves and more challenging samples are being queried to the oracle. 
	
	\section{Conclusion}
	\label{sec:conclusion}
	Image segmentation is one of the most challenging computer vision tasks, particularly for biomedical data such as electron microscopy as annotations are sparsely available. In order to be practically usable and scalable, image segmentation algorithms such as U-Net need to be able to cope with smaller amounts of annotated data. In this work, we propose to employ recent active learning approaches to minimize annotation efforts for training segmentation networks. Specifically, several of these approaches (\eg maximum entropy and least confidence sampling) obtain the same performance as the random sampling baseline, but require 4 times fewer annotations. In future work, we will further minimize labeling efforts, by combining this active learning paradigm with weakly supervised approaches (\ie using partially annotated data). 
	
	\bibliographystyle{splncs04}
	\bibliography{references}
\end{document}